\documentclass[letterpaper]{article} 
\usepackage[]{aaai2026}  
\usepackage{times}  
\usepackage{helvet}  
\usepackage{courier}  
\usepackage[hyphens]{url}  
\usepackage{graphicx} 
\usepackage{natbib}  
\usepackage{caption} 
\usepackage{algorithm}
\usepackage{algorithmic}
\usepackage{float}
\usepackage{subcaption}
\usepackage{multirow}
\usepackage{booktabs}
\usepackage{amsmath}
\usepackage{amssymb}
\usepackage[table]{xcolor}
\usepackage{tabularx}
\usepackage{bm}

\usepackage{newfloat}
\usepackage{listings}
\DeclareCaptionStyle{ruled}{labelfont=normalfont,labelsep=colon,strut=off} 
\floatstyle{ruled}
\newfloat{listing}{tb}{lst}{}
\floatname{listing}{Listing}
\title{SUGAR: Learning Skeleton Representation with Visual-Motion Knowledge for Action Recognition}
\author {
    Qilang Ye\textsuperscript{\rm 1} \textsuperscript{\rm 2}, 
    Yu Zhou\textsuperscript{\rm 1} \textsuperscript{\rm 2}\thanks{Corresponding Authors (yzhou@nankai.edu.cn, yuzitong@gbu.edu.cn)},
    Lian He\textsuperscript{\rm 2} \textsuperscript{\rm 5},
    Jie Zhang\textsuperscript{\rm 3}, 
    Xuanming Guo\textsuperscript{\rm 2}, 
    Jiayu Zhang\textsuperscript{\rm 3}, 
    Mingkui Tan\textsuperscript{\rm 6},
    Weicheng Xie\textsuperscript{\rm 7},
    Yue Sun\textsuperscript{\rm 8},
    Tao Tan\textsuperscript{\rm 8},
    Xiaochen Yuan\textsuperscript{\rm 8},
    Ghada Khoriba\textsuperscript{\rm 9},
    Zitong Yu\textsuperscript{\rm 3} \textsuperscript{\rm 4}\footnotemark[\value{footnote}]\\
}
\affiliations {
    \textsuperscript{\rm 1}VCIP \& TMCC \& DISSec, College of Computer Science \& College of Cryptology and Cyber Science, Nankai University\\
    \textsuperscript{\rm 2}Beijing Zhongguancun Academy\\
    \textsuperscript{\rm 3}Great Bay University\\
    \textsuperscript{\rm 4}Dongguan Key Laboratory for Intelligence and Information Technology\\
    \textsuperscript{\rm 5}School of Computer Science \& Technology, Beijing Institute of Technology\\
    \textsuperscript{\rm 6}South China University of Technology\\
    \textsuperscript{\rm 7}Shenzhen University
    \textsuperscript{\rm 8}Macao Polytechnic University
    \textsuperscript{\rm 9}Nile University\\
    yql@mail.nankai.edu.cn
}

\usepackage{bibentry}

\begin{document}

\maketitle

\begin{abstract}
Large Language Models (LLMs) hold rich implicit knowledge and powerful transferability. In this paper, we explore the combination of LLMs with the human skeleton to perform action classification and description. However, when treating LLM as a recognizer, two questions arise: 1) How can LLMs understand skeleton? 2) How can LLMs distinguish among actions? To address these problems, we introduce a novel paradigm named learning \textbf{\large S}keleton representation with vis\textbf{\large U}al-motion knowled\textbf{\large G}e for \textbf{\large A}ction \textbf{\large R}ecognition (SUGAR). In our pipeline, we first utilize off-the-shelf large-scale video models as a knowledge base to generate visual, motion information related to actions. Then, we propose to supervise skeleton learning through this prior knowledge to yield discrete representations. Finally, we use the LLM with untouched pre-training weights to understand these representations and generate the desired action targets and descriptions. Notably, we present a Temporal Query Projection (TQP) module to continuously model the skeleton signals with long sequences. Experiments on several skeleton-based action classification benchmarks demonstrate the efficacy of our SUGAR. Moreover, experiments on zero-shot scenarios show that SUGAR is more versatile than linear-based methods.
\end{abstract}
\section{Introduction}
\label{sec:intro}

\begin{figure}
	\centering
		\includegraphics[scale=0.36]{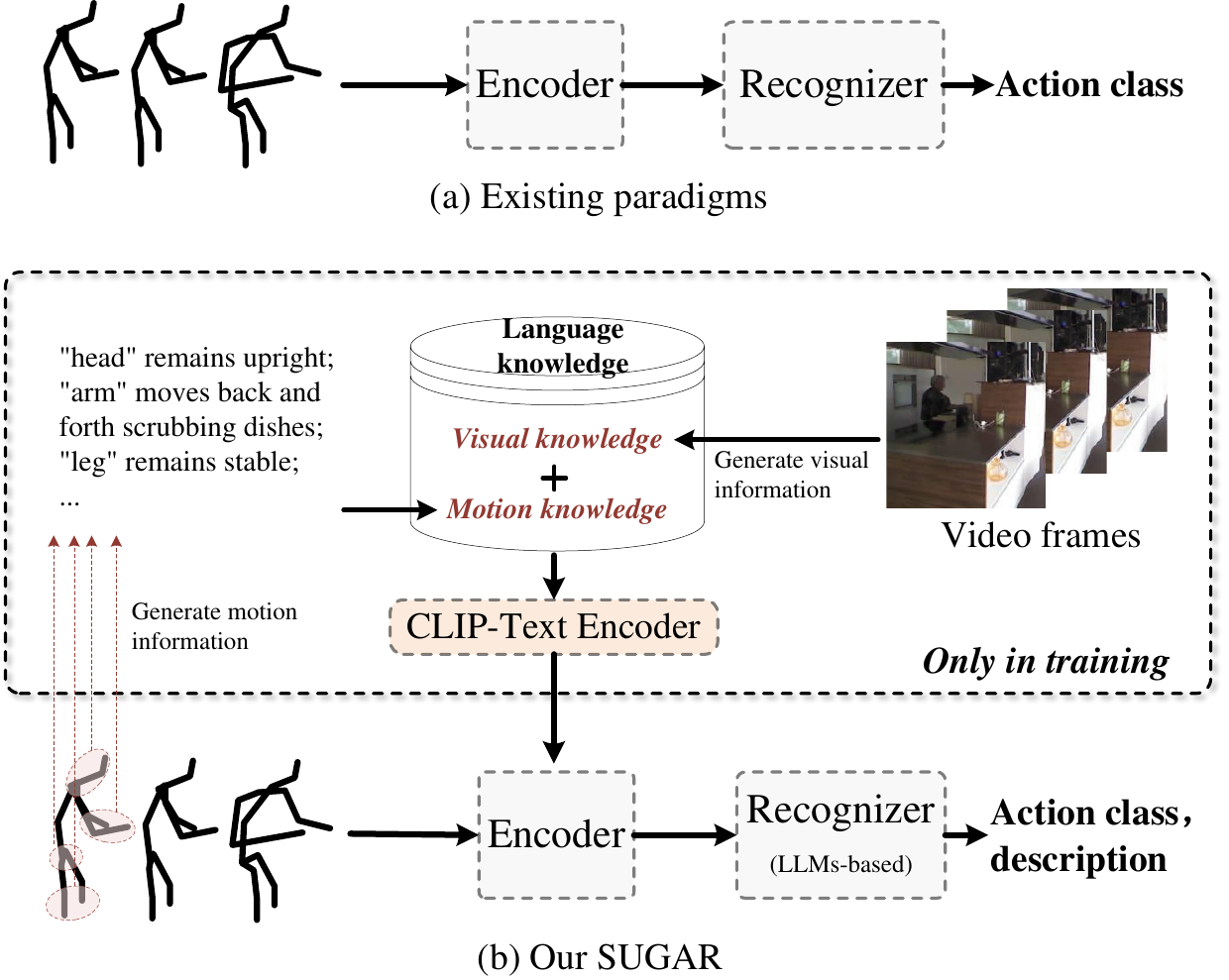}
\caption{Comparison between existing action recognition paradigms and our SUGAR.}
	\label{fig:fig1}
\end{figure}

Skeleton-based action recognition \cite{skl-2,2sAGCN,ctrgcn,ske2grid,skl-3} aims to model the spatio-temporal graph structure \cite{st-gcn} to classify actions. As representations of human behavior, skeletons satisfy practical applications such as human-computer interaction and intelligent monitoring with lightweight data storage. Although training with skeleton data (e.g. human body joints) brings computational efficiency \cite{lightweightSKL-1,lightweightSKL-2}, building an efficient recognizer for most activities of daily living is challenging and requires distinguishing similar fine-grained actions \cite{toyota}. For example, the movement trajectories of drinking water and snacking are extremely similar, leading the recognizer to easily categorize them as one.

Recently, Large Language Models (LLMs) such as Vicuna \cite{vicuna} and LLaMA \cite{llama2} have greatly influenced different domains. They can be effectively applied to many tasks that are not limited to plain text, but also to visual as well as audio tasks \cite{videollama,vtimellm,videochat}. A great deal of work has started to leverage the generated rich linguistic knowledge to improve their subtasks. LLMs can generate relevant information to help the corresponding network to learn linguistic knowledge and improve reasoning ability. Moreover, people have found that LLMs can be used not only as interpreters for text generation, but also as recognizers or classifiers to optimize downstream tasks \cite{CAT}. Specifically, fine-tuning on a large-scale multimodal corpus, LLMs can then adapt to different modal inputs. 

But \textit{what if the large language model served as an action recognizer?} Qu et al. \cite{actionllm} have given us answers. LLM contains rich human-centric knowledge after pre-training over a large corpus, it holds strong power to learn the input skeleton signal. Different from traditional linear-based methods \cite{st-gcn,2sAGCN}, LLMs are not only capable of classifying but also describing actions. However, several problems remain to be explored: \textbf{\textit{1) How can LLMs understand skeleton?}} LLMs can model language sequences and targets, but remain a challenge with input from other ``non-human languages''. There is a gap between the skeleton and the text, and a lack of a universal encoder to align the two. Previous works \cite{actionllm,LearningDiscriminativeRepresentations,LearningDiscriminativeRepresentations2,LearningDiscriminativeRepresentations3} introduce an action-based vector quantized variational autoencoder \cite{vqvae} to learn discrete tokens, it provides a solution to transfer the skeleton into the LLM. However, LLMs are mostly pre-training over human language datasets, so it is difficult to ensure that the skeleton tokens are consistent with text tokens. \textbf{\textit{2) How can LLMs distinguish between similar skeleton signals?}} As we mentioned, daily activities are filled with many similar actions. When the appearance information is missing, it is difficult to distinguish whether we are drinking or eating.

To solve the above problems, we aim to make similar actions in daily activities towards more discrete representations, and to make LLMs understand such representations for downstream tasks. Therefore, we propose a novel paradigm named learning \textbf{\large S}keleton representation with vis\textbf{\large U}al-motion
knowled\textbf{\large G}e for \textbf{\large A}ction \textbf{\large R}ecognition (SUGAR). As shown in Fig. \ref{fig:fig1}, most of the existing paradigms \cite{2sAGCN,st-gcn,ctrgcn}, including LLMs-based method \cite{actionllm}, only emphasize how to design powerful recognizers to achieve excellent performance and then output the action targets. In contrast, we develop a new training scheme, which leverages generated visual and motion information to supervise the learning process in skeletons. We believe that this a priori knowledge is crucial for LLMs as a recognizer.

Specifically, our SUGAR mainly targets the mentioned problems \textbf{\textit{1)}} and \textbf{\textit{2)}}. To make LLMs understand the skeleton input, we first define a series of action lists and fine-tune the LLMs with a fixed instruction. Specifically, we leverage the pre-trained skeleton encoder to generate representations, and then perform low-rank adaptation (LoRA) \cite{lora} on the LLMs to make the model understand the input skeleton representations. Notably, we design a Temporal Query Projection (TQP) module to shorten the input representation sequence and maintain continuous temporal modeling.

To better distinguish similar fine-grained actions, we propose to introduce motion and visual information \footnote{Motion and visual information refer to the text knowledge extracted from the motion skeleton and the video, respectively.}. During movement, each part of each action has a different trajectory, and describing this trajectory can lead to greater variability between different movements. Moreover, such high-level semantic information can provide fine-grained prior knowledge for skeleton representation learning. However, when motion trajectories are exactly similar (e.g., eating and drinking), motion information alone is not sufficient. Therefore, we use an off-the-shelf Visual Language Model (VLM) \cite{gpt4} to automate the synthesis of visual information related to actions. Finally, we store visual and motion information as text and learn discrete skeleton representations in a way that mimics the CLIP-based pre-training of Image-Text \cite{clip}. In this way, LLMs can learn a skeleton representation that is aligned with the corresponding description.

In summary, our main contributions of this work include:
\begin{enumerate}
\item Different from previous skeleton-only paradigms, we introduce visual and motion knowledge to learn a more discrete skeleton representation. Such representations can allow the recognizer to better distinguish between different similar actions.
\item We propose a novel LLMs-based paradigm, namely SUGAR, for action recognition. It leverages LLM's untouched pre-trained weights to classify, and includes a Temporal Query Projection (TQP) module to reduce computational cost while enhancing long-term skeleton dynamic modeling.
\item We achieve state-of-the-art performance on several skeleton-based action classification benchmarks. One highlight is that we apply this paradigm to zero-shot scenarios and the experimental results show that SUGAR has strong generalization under zero-shot scenarios, which demonstrates the advantages of LLMs as an alternative to linear-based methods.
\end{enumerate}

\begin{figure*}
	\centering
		\includegraphics[scale=0.44]{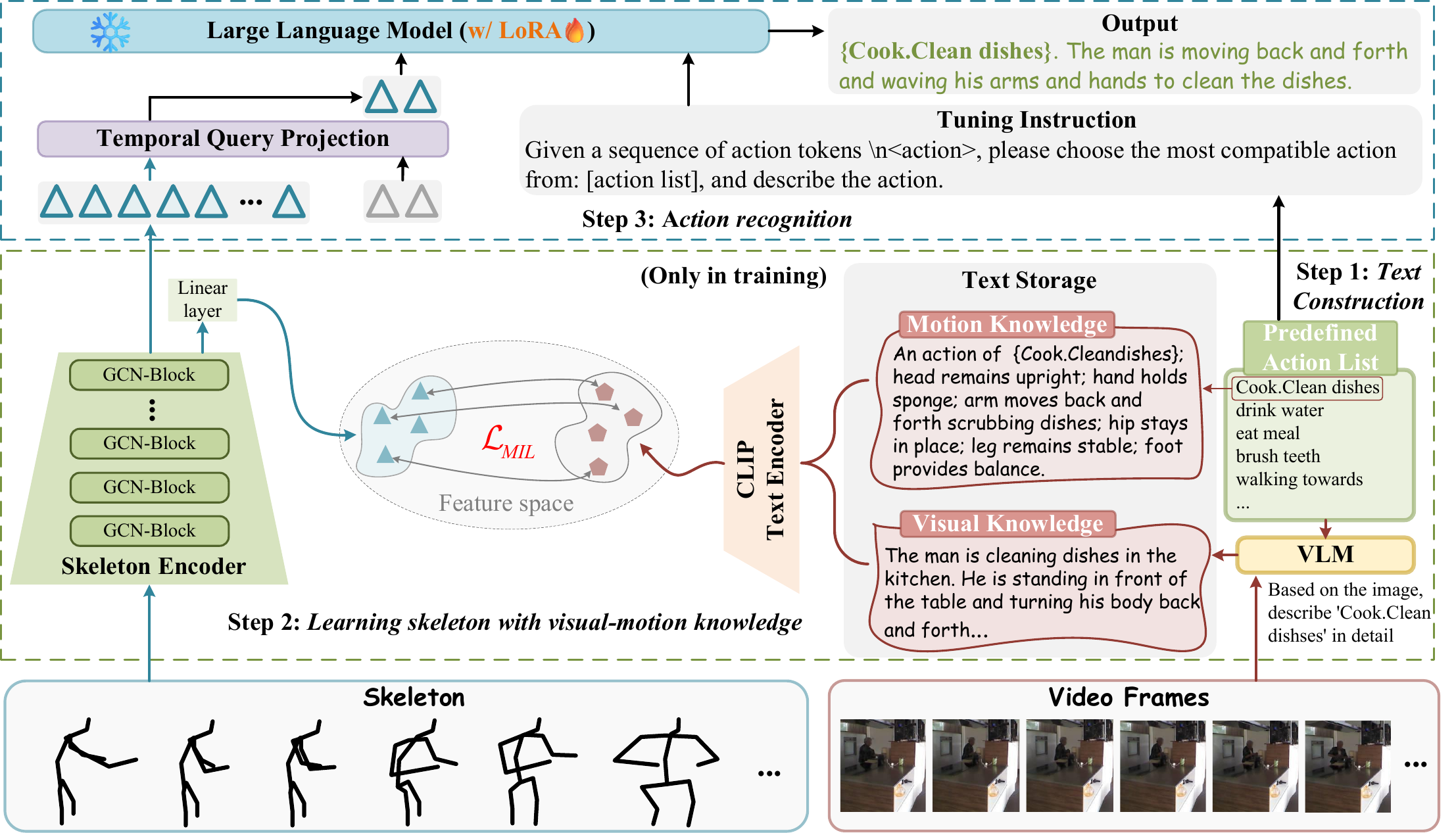}
\caption{Overall framework of SUGAR. The complete training procedure is divided into three parts. We use the GPT-generated fine-grained action description and VLM-generated visual description as input for the text encoder to supervise the skeleton representation learning, where the linear layer maps the skeleton to the same feature space as the text. During inference, only the skeleton data needs to be input for action recognition.}
	\label{fig:framework}
\end{figure*}

\section{Related Work}

\noindent \textbf{Unimodal and Multimodal Action Recognition.} Unimodal methods for action recognition can be categorized into RGB-based and skeleton-based (joint+bone), both of which perform inference from a single modality of information. RGB-based methods \cite{ye2024pose,nan20243sg,rajendran2024review,li2025role} mainly deal with a sequence of image information, and are good at capturing rich visual contexts. However, they usually suffer from excessive parameters and overfitting. On the other hand, the skeleton-based recognition performance has made great progress from the ST-GCN model proposed by Yan et al. \cite{st-gcn}. Afterwards, models like 2s-AGCN \cite{2sAGCN} and CTR-GCN \cite{ctrgcn} use dual-stream to dynamically learn the topology and achieve higher accuracy. However, these methods have the drawback of lacking visual information support and depending on accurate human pose estimation techniques.

The combination of skeleton and visual can enhance the recognition of actions. Fabien et al. \cite{hands} proposed a novel multimodal approach that focuses on the hand pose. This method opens up new possibilities for future work but it ignores some details of the rest of the body. Some works \cite{toyota} attempt to weight fusion of RGB images by utilizing pose data, which is an innovative way of integrating different types of modality. Other works \cite{vpn, mmnet} introduced a spatial embedding to map pose data to visual features. \cite{vpn++} developed a feature-level and attention-level distillation, which offers a practical solution for combining RGB and pose. However, the fusion of multimodal inputs requires massive computation. Moreover, we have no way to know whether heterogeneous visualizations and skeletons can be effectively combined in the learning process.

\noindent \textbf{Vision-Language Models for Action Recognition.} Co-training multiple modalities can learn powerful representations for downstream tasks. For example, vision-language pre-trained models such as CLIP \cite{clip}, BLIP \cite{blip}, and ALIGN \cite{align} have demonstrated that different modalities can be approximated between the two representations by defined learning objectives. Besides the contrastive learning of images and language, some work has started to apply representation learning to the action recognition domain. Wang et al. \cite{actionclip} follow the CLIP training strategy to help with downstream action recognition tasks. They convert action labels into representations and perform representation learning by calculating the similarity with the video. However, we argue that labels like ``\textit{A action of \{\}}'' lack substantial semantic information and do not provide exhaustive knowledge. Other efforts \cite{xie2024fusionmamba,lin2025reliable} to introduce representation learning into skeleton still do not provide rich linguistic knowledge to aid learning. To learn a more discrete representation, we introduce linguistic knowledge such as motion and vision to improve the performance.

\noindent \textbf{Large Language Models for Action Recognition.} A variety of studies \cite{CAT,ye2025cat+, videochat} have shown that the powerful generalization ability of LLMs can help different downstream task reasoning. With the help of prompt learning \cite{promptlearning1}, LLMs can generate any form of text with any content. Recently, some work has started to focus on how to apply LLMs to action-centric tasks. Qu et al. \cite{actionllm} use VQ-VAE \cite{vqvae} to learn specific action tokens and fine-tune large models using LoRA \cite{lora}. Drawing on their ideas, we find that utilizing rich linguistic knowledge to learn skeleton representations can bring better performance to the recogniser and performs well in zero-shot scenarios.

\section{Methods}

\subsection{Overview of SUGAR}
As illustrated in Fig. \ref{fig:framework}, the workflow of SUGAR starts with constructing sets of texts related to action and vision. We collect a detailed description of each action and related visual information through a predefined list of actions (a dictionary that contains possible action labels). To avoid manual annotation, we have downloaded the powerful visual language model and GPT as generative tools. Then, we aim to pre-train a robust skeleton encoder that can be aligned to the text. Given a skeleton sequence as the input, the goal of the encoder is to make LLMs learn a more discrete skeleton representation. After that, we define a fixed instruction and fine-tune the LLM to understand the input skeleton representation. During inference, we only need to input the skeleton sequence to perform the action recognition task.
\subsection{Step 1: Text Construction}\label{sec1}
\noindent \textbf{Predefined Action List.}
We usually expect an ideal recognizer to be generalizable to different practical scenarios. Although existing LLMs are capable of open-ended answers, it remains a challenge to answer specific classes in the field of action recognition accurately. Therefore, we define a reasonable action dictionary, which includes all possible category names of the action dataset we used. Furthermore, we test our SUGAR in Sec. \ref{sec2} to show that: without fine-tuning, the model is able to achieve zero-shot inference with the predefined action list.

\noindent \textbf{Generate Motion Knowledge.}
We find that each action has a corresponding unique description, such information can be used as vectors to supervise the learning of each action set. However, it is difficult to make an action specific when using instance-level description. Therefore, we utilize a fine-grained human body part pattern to represent an action. For example, an action of ``Drink'' can be represented as the head tilting back slightly, the hand grasping the cup, and so on. 

Specifically, we follow the production process of the HAKE dataset \cite{HAKEA2V} and resolve each action into six body part movements: head, hand, arm, hip, leg, and foot. Moreover, to avoid manual annotation, we use the large language model GPT 3.5 turbo for action text generation. Given an action label from predefined action lists, we design the following prompts to generate motion descriptions $\mathcal{T}_{m}$:
\begin{figure}[!h]
	\centering
		\includegraphics[scale=0.6]{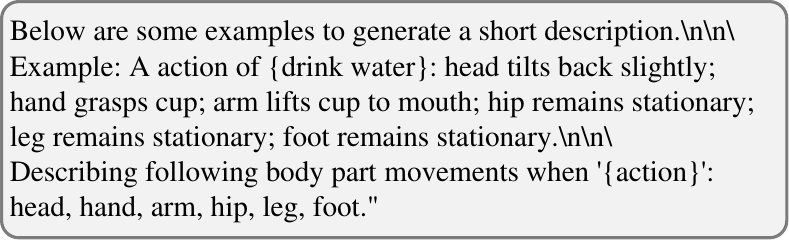}
\end{figure}

\noindent With our human-written examples, the large-scale model can generalize them and produce the desired action description without any parameter updates.

\noindent \textbf{Generate Action-related Visual Knowledge.}
We employ a powerful off-the-shelf vision-language model GPT-4V \cite{gpt4} for generating action-related visual information. However, LLMs' responses often contain redundancy and nonsense, which is not conducive to represent a particular class. For example, the VLM is inclined to generate a person's appearance characteristics, the environment, etc. Therefore, we establish three rules to limit the generation of VLMs: - \textit{Provides a description of the given image that correctly matches the action}, - \textit{Only describe the scene related to the given action}, - \textit{Do not provide any text or explanations unrelated to the action}.

Indoor activities generally involve long static movements and repetitive frames. To filter out the large number of duplicates in the generated frame descriptions, we transform each frame via the vision encoder from CLIP into features and collect the most dissimilar image set by counting the similarity scores between them. Then we perform text generation for each frame from the collected image set and obtain the visual description set $\mathcal{T}_{v}$.


\subsection{Step 2: Skeleton Representation Learning}\label{sec2}
\noindent \textbf{Skeleton Encoder.}
The skeleton is stored as coordinates, it constructs complex and diverse graphs with joint connections. Graph Convolution Network (GCN) \cite{gcn} can be applied to graph signal processing due to its ability to aggregate node information. In this paper, we stack multiple GCN blocks as a skeleton encoder, and each block performs information aggregation on the input to learn the representation. Specifically, the input skeleton can be denoted as $G=\{V,\mathcal{E}\}$, a graph that contains $V$ human joints and $\mathcal{E}$ is the set of edges. We define the aggregated features of the input skeleton at layer $l$ as $\text{H}^{l} \in {\mathbb{R}^{D \times F}}$, the $F$ denotes the feature dimension. The graph convolution can be represented as:
\begin{equation}
\text{H}^{l+1} = \sigma(\text{D}^{-{\frac{1}{2}}} \text{A} \text{D}^{-{\frac{1}{2}}} \text{H}^{l} \text{W}^{l}),
\end{equation}
where $\text{D} \in {\mathbb{R}^{D \times D}}$ is the degree matrix, $\text{A}$ is the adjacency matrix of graph $G$, $\text{W}^{l}$ is the $l$-layer's weight parameter, and $\sigma$ is the activation function.

Besides spatial aggregation, we leverage the multi-scale temporal modeling module designed by \cite{ctrgcn} to model the action in the temporal dimension. However, Maxpool in the original method loses fine-grained temporal information. We note that skeleton temporal sequences are critical for fine-grained action inference. Therefore, we discard the pooling operation in the time dimension and keep the complete time information for the next step.

\noindent \textbf{Text Encoder.}
We employ a pre-trained CLIP-based \cite{clip} text encoder $E_{t}()$ to transfer the collected information into embedding. Notably, we consider that motion description and visual description belong to two different categories, so we divide the output into two parts as follows:
\begin{equation}
m = E_{t}(\mathcal{T}_{m}), \quad v_i =E_{t}(\mathcal{T}_{v_{i}}),
\end{equation}
where $i \in I_v$, $I_{v}$ are indices of the collected visual knowledge. To diversify the collected linguistic knowledge, we randomly combine the motion description embedding and the visual description embedding to obtain $\mathbf{t}=\{m,v_i|i \in I_v\}$.

\noindent \textbf{Contrastive Learning for Skeleton and Text.}
Contrary to one-to-one contrastive learning, we emphasize that the skeleton representation can be positively matched to multiple texts. Inspired by \cite{loss_mil} that combines Multiple Instance Learning (MIL) and Noise Contrastive Estimation for contrastive learning, we further propose to encourage the learning of skeleton representations under multiple texts supervision. Specifically, the optimal goal is to contrast skeleton-text within the batch:
\begin{equation}
\mathcal{L}_{MIL} = - \frac{1}{{\left| B \right|}}\sum\limits_i {\log \frac{{\sum\nolimits_j {\sum\nolimits_n {\exp ({\mathbf{s}_i}^ \top {\mathbf{t}_{j,n}}/\tau )} } }}{{\sum\nolimits_k {\sum\nolimits_n {\exp ({\mathbf{{s}_i}^ \top {\mathbf{t}_{k,n}}/\tau )} } }}}},
\end{equation}
where $\mathbf{s}$ is the encoded feature of the skeleton, $B$ is the batch size, $i,j,k \in B$, $n$ is the index of the corresponding visual-motion description embedding in $\mathbf{t}$. $\tau$ is the temperature parameter.

\subsection{Step 3: Action Recognition}\label{sec3}
\noindent \textbf{Temporal Query Projection.}
To map the skeleton representation into the LLMs' embedding space, we have to design a multimodal projector that converts the embedding inputs into language tokens of a suitable length. Earlier projectors for processing long sequence embeddings \cite{Valley,VLM-Eval} usually perform temporal compression or employ a linear layer to map to the corresponding dimension. However, we consider that the temporal signals of the skeleton are constructed from a continuous series of positional information, and that pooling or arbitrary extraction would disrupt the context of the entire topology. Therefore, we introduce a novel projection method named Temporal Query Projection (TQP), which can continuously query temporal signal representations and distill them into short language tokens.
\begin{figure}
	\centering
		\includegraphics[scale=0.6]{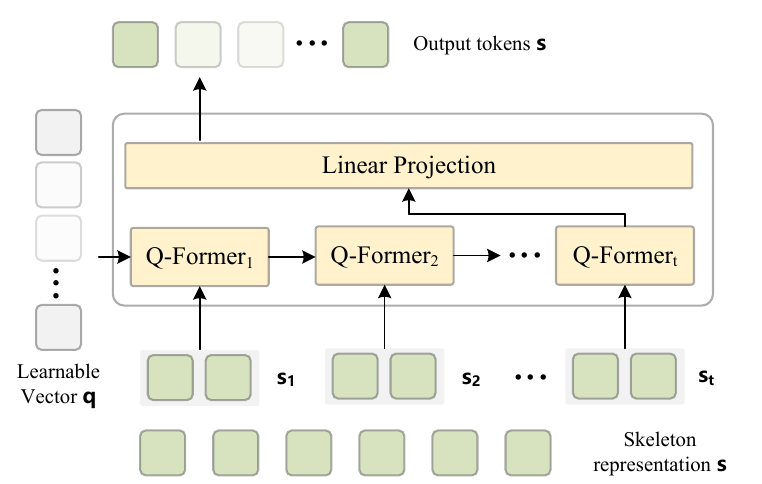}
\caption{Temporal Query Projection consists of a number of Q-Formers \cite{blip2} and a linear layer.}
	\label{fig:tqp}
\end{figure}


As shown in Fig. \ref{fig:tqp}, let $\mathbf{s}\in {\mathbb{R}^{L_{s} \times d}}$ be the skeleton representation of length $L_{s}$. We introduce the Q-Former function $f_{Q}$ \cite{blip2} to replace the previous direct projection and cross-attention. Notably, we set a hyperparameter $k$ to select $t$ time points for a skeleton segment $\{ \mathbf{s}_{i} | i=1,2,...,t \}$, where $t=\frac{L_{s}}{k}$. Then we customize a learnable query vector $\mathbf{q} \in {\mathbb{R}^{k \times d}}$ in length $k$. Contrary to a single Q-Former, we utilize the previously queried skeleton representation as the next query in order to continuously model the entire skeleton signal, which can be denoted as:
\begin{equation}
\hat{\mathbf{s}}_{t} = {f_{Q}}^{t}(\mathbf{s}_{t-1},\mathbf{s}_{t}),
\end{equation}
where ${f_{Q}}^{t}$ denotes $t$-th $f_{Q}$, $\hat{\mathbf{s}}_{t} \in {\mathbb{R}^{L \times d}}$, and all the parameters of Q-Former are shared.

\noindent \textbf{Training Strategy.}
The training of SUGAR is mainly divided into two stages. The first stage is training a skeleton encoder supervised by visual-motion knowledge with the learning objective $\mathcal{L}_{MIL}$. The second stage is to fine-tune the LLMs with LoRA to adapt the non-human language tokens with the learning objective $\mathcal{L}_{Lora}$. Specifically, we define the LLM recognizer as $f_{LLM}()$, during the fine-tuning phase, we input the instructions shown in Fig. \ref{fig:framework} and the action tokens $\hat{\mathbf{s}}$ after TQP into the LLM to predict the action class. $\mathcal{L}_{Lora}$ can be expressed as:
\begin{equation}
\mathcal{L}_{Lora} = CrossEntropy(f_{LLM}(\hat{\mathbf{s}}),y),
\end{equation}
where we use cross-entropy loss to supervise the fine-tuning of the LLM, and $y$ are tokens containing the corresponding ground truth and a description of the action. Furthermore, to make the output of the LLM more than just a single category, we collect a small number of brief descriptions of actions as instructions to fine-tune the model, where we use GPT to generate such brief descriptions based on the collected motion knowledge.

\noindent \textbf{Inference.} During inference, we only need to input the skeleton signal, and then use the pre-trained language-enhanced skeleton encoder to transform them into a discrete skeleton representation. After the projection process with TQP, LLMs predict the most appropriate one based on the predefined list of actions and output a brief description of the action.

\begin{table*}[]

\centering
\label{exp1}
\setlength{\tabcolsep}{0.8mm}{
\begin{tabular}{lccccccccccc}
\toprule  
\multirow{2}{*}{Methods} & \multirow{2}{*}{Recognizer} & \multirow{2}{*}{Skeleton} &\multicolumn{3}{c}{Toyota Smarthome}&\multicolumn{2}{c}{PKU-MMD} &\multicolumn{2}{c}{NTU RGB+D}&\multicolumn{2}{c}{NTU RGB+D 120}\\
\cmidrule(lr){4-6}\cmidrule(lr){7-8}\cmidrule(lr){9-10}\cmidrule(lr){11-12}
 &&& $X\text{-}sub$&$X\text{-}view1$&$X\text{-}view2$&$X\text{-}sub$&$X\text{-}view$&$X\text{-}sub$&$X\text{-}view$&$X\text{-}sub$&$X\text{-}view$ \\
\midrule  
2s-AGCN & FC & Joint + Bone & 55.7 & 21.6 & 53.3& 84.5 & 92.1& 84.2 & 93.0& 78.2& 82.9\\
MS-G3D & FC & Joint + Bone & 55.9 & 17.4 & 56.7& - & -& 86.0 & 94.1& 80.2& 86.1\\
ST-GCN & FC & Joint + Bone & 62.9 & 40.6 & 51.4 & 86.7 & 92.7& 81.5 & 88.3 &82.1 & 84.5\\
Shift-GCN  & FC & Joint + Bone& - & - & - & - & -& 87.8 & 95.1& 80.9 & 83.2\\
SSTA-PRS  & FC & Joint + Bone& 62.1 & 22.8 & 54.0 & - & -& -& - & - & -\\
ML-STGNet & FC & Joint + Bone& 64.6 & 29.9 &63.5& - & -& -& - & - & -\\
UNIK & FC & Joint + Bone& 62.1 & 33.4 & 63.6& - & -& 86.8 & 94.4 & 80.8& 86.5 \\
LLM-AR & LLM& Joint& 67.0 & 36.1 & 66.6& - & -& 95.0 & \textbf{98.4} & 88.7& \textbf{91.5} \\
\textbf{SUGAR} (Ours) & LLM& Joint& \textbf{70.2} & \textbf{50.9} & \textbf{67.1}& \textbf{89.0} & \textbf{94.3} & \textbf{95.2} & \underline{97.8}& \textbf{90.1} & \underline{89.7}  \\
\bottomrule  
\end{tabular}}
\caption{Comparison result (\%) of skeleton-based methods on different action classification benchmarks. FC denotes linear-based methods, and LLM denotes LLMs-based methods.}
\end{table*}

\section{Experiments}
\subsection{Datasets} 
\noindent \textbf{Toyota Smarthome} (SH) \cite{toyota} dataset contains 31 activities and generates a total of 16,115 video samples. The official proposal to get the skeleton sequence by \cite{toyotaskl}, and the dataset provide three evaluation protocols: cross-subject ($X\text{-}sub$) and cross-view ($X\text{-}view1$ and $X\text{-}view2$). Results on Toyota Smarthome are mean per-class accuracy.

\noindent \textbf{NTU RGB+D} (NTU 60) \cite{ntu60} contains 60 action categories with 56,880 video samples, and it provides the original skeleton sequences. This dataset provides two evaluation protocols: cross-subject ($X\text{-}sub$) and cross-view ($X\text{-}view$).

\noindent \textbf{NTU RGB+D} (NTU 120) \cite{ntu120} is an upgraded version of NTU 60, it contains more than 114k skeleton sequences and it has 120 action categories with 57,600 video samples. This dataset provides two evaluation protocols: cross-subject ($X\text{-}sub$) and cross-view ($X\text{-}view$).

\noindent \textbf{PKU-MMD} (PKU) \cite{pku} contains a total of 21,545 samples with skeleton data and 51 action categories. This dataset provides two evaluation protocols: cross-subject ($X\text{-}sub$) and cross-view ($X\text{-}view$).

\subsection{Implementation Details}
For Step 1, we use the labels of the corresponding dataset as the predefined action list, and we use GPT 3.5 turbo for action text generation and GPT-4V \cite{gpt4} for action-related visual text generation. For Step 2, the skeleton data is pre-processed by the code of \cite{skl_code}. We use CTR-GCN \cite{ctrgcn} as the skeleton encoder backbone and CLIP \cite{clip} as the text encoder. Then, we train the skeleton encoder using the SGD optimizer with an initial learning rate of 0.01 for a total number of 200 epochs with batch size 200 (100 for Toyota Smarthome) and reduced by a factor of 0.1. For Step 3, we freeze all parameters of the skeleton encoder and we use the LLaMA2 7B \cite{llama2} as the LLM. We set r = 64 and alpha = 16 for the LoRA parameters, and the total batch size is set to 128 for training 1 epoch with a learning rate of $2e^{-5}$. All experiments are conducted on two NVIDIA A6000 GPUs.

\subsection{Comparisons to the State of the Arts}
We compare our SUGAR with previous linear-based methods and LLMs-based methods in Table \ref{exp1}. Different from most methods that use the ensemble strategies (Joint with Motion, Bone with Motion, Joint with Bone), we train the entire framework using only the joints of the skeleton data as inputs, in a way that drastically reduces the computational cost. It is clear that our SUGAR dramatically outperforms all state-of-the-art results of linear-based classification methods in the Toyota Smarthome dataset \cite{toyota}. Comparing to the same type of LLMs-based method, LLM-AR \cite{actionllm}, we outperform on $X\text{-}sub$ and $X\text{-}view1$ by $3.2\%$ and $14.8\%$, respectively. We analyze that: the activities recorded in the Toyota Smarthome cannot be distinguished solely by their motion trajectories. Benefiting from the introduction of visual knowledge, the skeleton can learn more diverse representations from high-level visual semantics, which makes it easier to recognize such composite actions compared to skeleton-only methods.

For the results in NTU RGB+D \cite{ntu60}, NTU RGB+D 120 \cite{ntu120}, and PKU-MMD \cite{pku}, such datasets record actions in a single scene. Compared to previous state-of-the-art methods, we consistently achieve competitive results across all the evaluation protocols. Furthermore, our paradigm of introducing motion knowledge to supervise the learning of skeletons makes the LLM obtain a more discrete representation. We strongly believe that this way has greater generality in action recognition, especially in zero-shot reasoning (which is discussed in Sec. \ref{exp-zero}).

\begin{table*}[!t]
	\centering
 \begin{minipage}{0.3\textwidth}
		\centering
		\small
		\makeatletter\def\@captype{table}\makeatother
		\label{tab-2}
        \setlength\tabcolsep{2pt}
		 \begin{tabular}{lcccc}
            \toprule  
            \multirow{2}{*}{Methods} & \multicolumn{2}{c}{Protocol 1} & \multicolumn{2}{c}{Protocol 2} \\
            &Top-1 & Top-5 &Top-1 & Top-5\\
            \midrule  
            \rowcolor{gray!30}\multicolumn{5}{c}{\textit{\footnotesize Linear-based}} \\
             ST-GCN & 30.1 & 45.2 & 36.9 & 55.2\\ 
             2S-AGCN  & 32.3 & 45.5 & 35.6 & 54.5\\ 
             CTR-GCN & 34.5 & 46.7 & 34.2 & 55.1\\ 
              \rowcolor{gray!30}\multicolumn{5}{c}{\textit{\footnotesize LLM-based}} \\
            LLM-AR & 59.7 & 84.1 & 49.4 & 74.2\\ 
            \textbf{SUGAR} & \textbf{65.3} & \textbf{89.8} & \textbf{53.4} & \textbf{77.6}\\ 
            \bottomrule  
            \end{tabular}
            \caption{Evaluation on unseen activities. We report the Top-1 and Top-5 overall accuracy.} 
	\end{minipage}\quad
        \begin{minipage}{0.3\textwidth}
		\centering
		\small
		\makeatletter\def\@captype{table}\makeatother
		 \begin{tabular}{lc}
            \toprule
            Language Knowledge & Acc(\%) \\
            \midrule
             w/o visual, motion & 69.2\\
             w/ visual& 69.4\\
             w/ motion& 72.1\\
            w/ visual, motion & \textbf{73.4}\\
            \bottomrule
        \end{tabular}
        \caption{Results on the overall accuracy of incorporating different language knowledge in SH.}\label{tab-3}
	\end{minipage}\quad
	\begin{minipage}{0.3\textwidth}
		\centering
		\small
		\makeatletter\def\@captype{table}\makeatother
		\begin{tabular}{lc}
            \hline  
            Methods & Acc (\%) \\
            \hline  
            X-Attn. & 52.1\\
            One Q-Former & 70.7\\
            One linear layer & 70.4\\
            Temporal query projection & \textbf{73.4} \\
            \hline  
            \end{tabular}
            \caption{Results on the overall accuracy of different bridging modules in SH \cite{toyota}. X-Attn. denotes mimicking cross-attention in \cite{CAT}.}\label{tab-5}
	\end{minipage}

\end{table*}

\subsection{Zero-shot on Action Recognition} \label{exp-zero}
Compared to traditional linear-based recognition approaches, we argue that the rich prior knowledge contained in large language models might benefit models' generalization capacities for recognizing unseen activities. To verify this conjecture, we build two evaluation protocols: the first is to pre-train on NTU 60 and then infer on 10 unseen action classes of NTU 120 cross-subject. The second one is cross-dataset testing, i.e., pre-training on NTU 60 and then inference on the test set of PKU-MMD cross-subject. All evaluation protocols report overall accuracy, and to explore performance in more detail, we instruct SUGAR to output what it considers to be the top five most compatible action categories as: ``Given a sequence of action tokens [action], please choose the top five most compatible action from [action list].''

To better represent the superiority of SUGAR for zero-shot inference, we design a linear-based classifier with ST-GCN \cite{st-gcn} as the backbone and pre-train it on NTU 60. As shown in Table \ref{tab-2}, we find an interesting phenomenon: SUGAR is highly adaptable to both unseen action categories and even action across datasets. Compared to linear-based methods, the logistic distribution of the classifier output cannot be directly adapted to action categories from other datasets, which is a limitation compared to LLMs-based methods. In summary, SUGAR holds a promising recognition ability for unseen activities, and this experiment justifies the need for large language models as an alternative to linear-based approaches.

\subsection{Ablation Studies}

\noindent \textbf{Impact of Visual-Motion Knowledge.}
The key to SUGAR is the injected visual-motion knowledge. To verify whether this a priori knowledge contributes to the learning of skeleton representations, we test the impact of the encoder on the classification results before and after adding this knowledge in Table \ref{tab-3}. Notably, we train a skeleton encoder without adding visual knowledge or motion knowledge using the cross-entropy loss. The results show that using language prior knowledge to guide encoder learning leads to a stable performance improvement of 4.2\% in overall accuracy.

\noindent \textbf{Impact of TQP.}
To validate the effectiveness of our core module TQP, we report in Table \ref{tab-5} the impact of comparing different bridging strategies on the Toyota Smarthome. Initially, we attempt to transform skeleton representations into action tokens using the three currently dominant bridging modules: linear layer, cross-attention, and Q-Former. As the results show, we are surprised that one linear layer can bring results that are already able to approach the state-of-the-art classifier-based methods. However, such a way poses the problem of generating excessively long tokens, resulting in a longer overall training time. On the contrary, Q-Former's query mechanism can effectively reduce the length of input tokens to minimize the computational cost. Further, our proposed TQP can query the temporal information of skeleton representation in a sequential manner, which improves our performance by 2.7\% compared to a single Q-Former.

\begin{figure}
	\centering

		\includegraphics[scale=0.35]{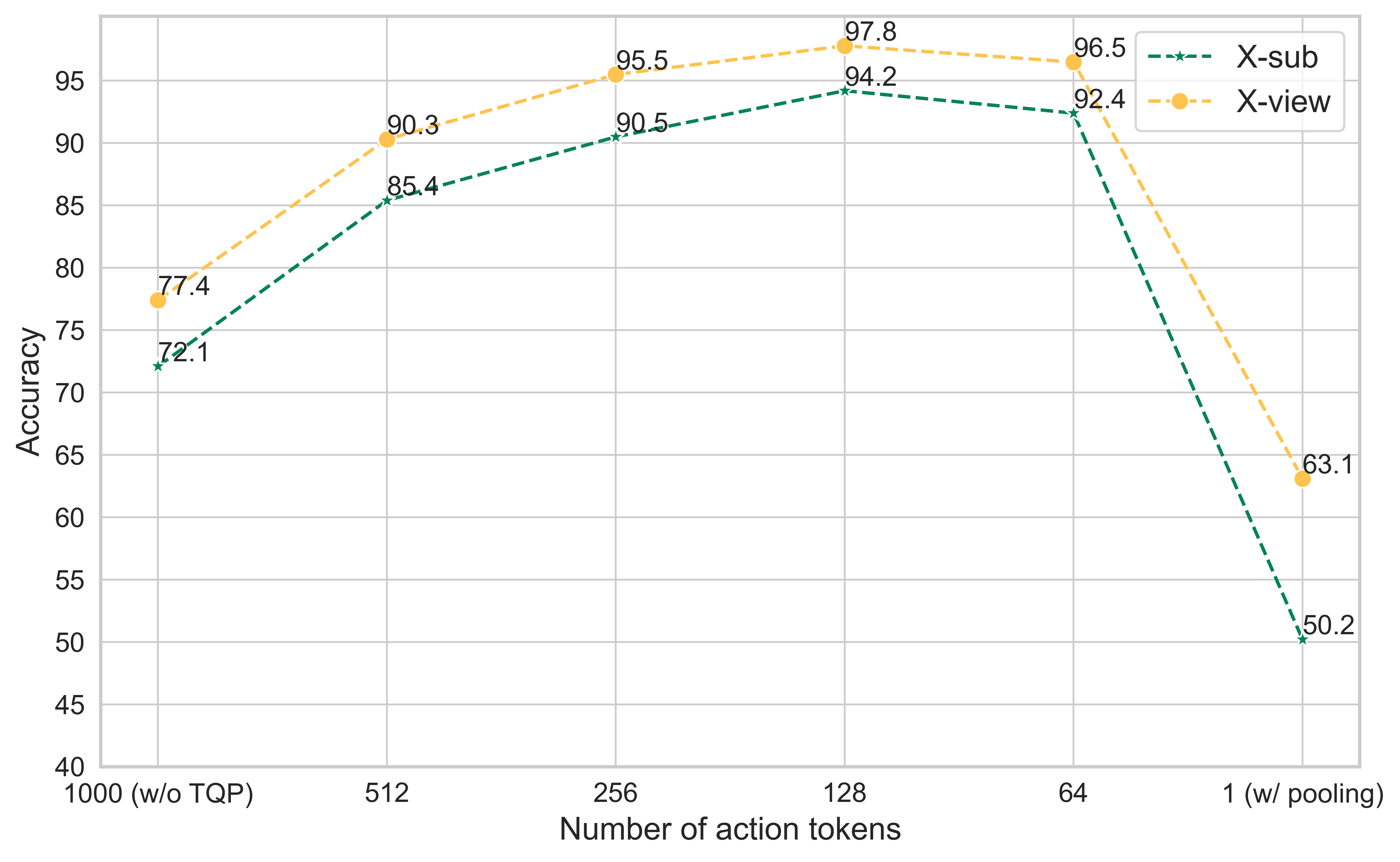}
\caption{The effect of the length of action tokens on NTU RGB+D, where 1000 denotes the length of the complete time series output by the skeleton encoder and 1 represents action tokens after temporal compression.}
	\label{fig:tokens}
\end{figure}

\noindent \textbf{Length of Action Tokens.}
Another matter worth exploring is the length of the action tokens that feed into the LLM. We evaluate the impact of different input lengths on NTU 60 in Fig. \ref{fig:tokens}. In this study, We test 6 different input lengths: 1000 (full length), 512, 256, 128, 64 and 1 (time-compressed) to find the most suitable one. The results show that by continuously reducing the input length, SUGAR shows signs of incremental increase. This further suggests that LLM may not excel at modeling non-language tokens with long temporal information. Furthermore, an interesting finding is that SUGAR has the worst results when we try to compress the length of action tokens to 1. We argue that many of the skeleton signals that represent different actions are too similar, and that performing compression makes such data more homogeneous to each other, leading to LLM's inability to discriminate between these actions. Therefore, we define the length $L$ of the query vector $\mathbf{q}$ in TQP as 128.

\begin{figure}[!h]
	\centering
	\begin{minipage}[c]{0.23\textwidth}
		\centering
		\includegraphics[width=\textwidth]{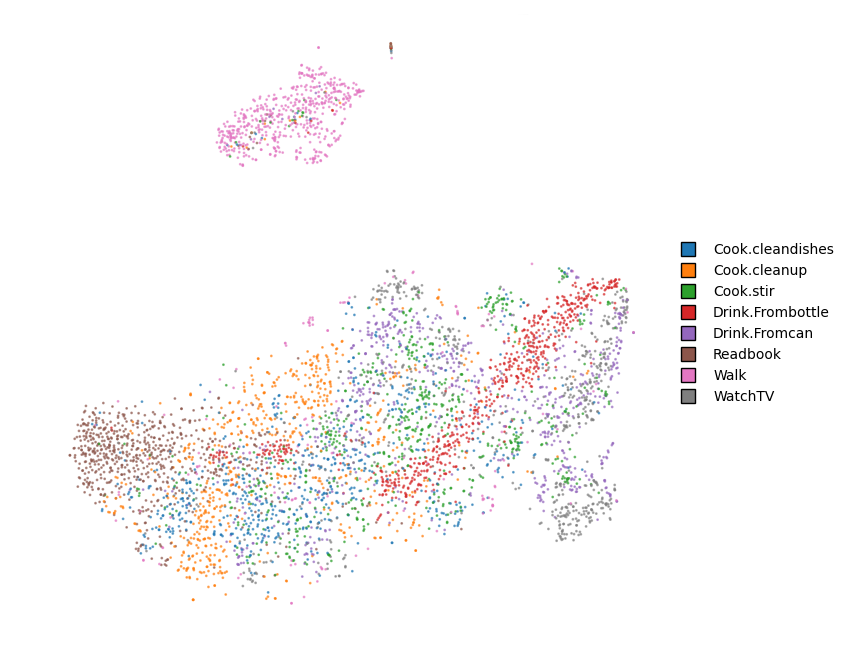}
		\subcaption{Before training}
		\label{fig_E2_2}
	\end{minipage} 
	\begin{minipage}[c]{0.23\textwidth}
		\centering
		\includegraphics[width=\textwidth]{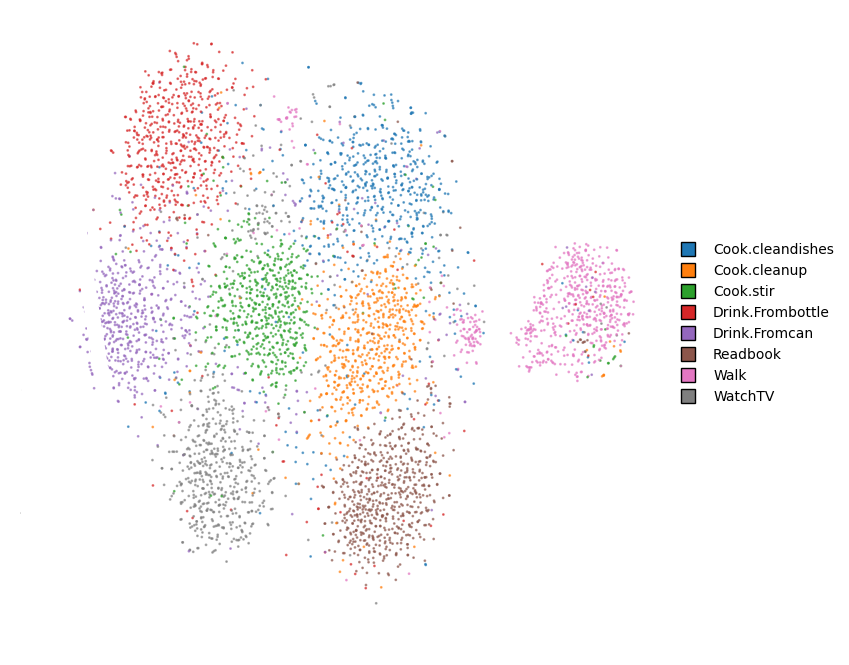}
		\subcaption{After training}
		\label{fig_E2_3}
	\end{minipage}
	\caption{t-SNE \cite{tsne} visualization of the learned skeleton representation. We show 8 action classes of the Toyota Smarthome dataset, indicated by colors.}
	\label{fig_tsne}
\end{figure}

\subsection{Visualization of Learned Representations}
We present the t-SNE \cite{tsne} visualization of the skeleton representation before and after the supervised training with language knowledge in Fig. \ref{fig_tsne}. Notably, we select visual distributions of similar action categories, e.g., ``Drink From bottle'' and ``Drink From can'', ``clean up'', and ``clean dishes''. We find that the representations acquired by introducing visual-motion knowledge to supervise during training are more discrete compared to the initial, and even similar actions are largely separated. We believe that in this way it allows LLM to better discriminate between similar actions.


\section{Conclusion}

In this paper, we develop a novel paradigm named SUGAR for skeleton-based action recognition, which introduces rich visual-motion knowledge to supervise the learning of skeleton representations and utilizes LLM as a recognizer. Experiments demonstrate that SUGAR is superior to traditional linear-based methods, especially under zero-shot scenarios, where SUGAR has a strong generalization capacity.

\section{Acknowledgments}
This research was funded by the Beijing Zhongguancun Academy (Grant No. 20240306), the National Natural Science Foundation of China (Grant No. 62376266 and 62406318), the CCF-Tencent Rhino-Bird Open Research Fund, Guangdong Research Team for Communication and Sensing Integrated with Intelligent Computing (Project No. 2024KCXTD047), SongShan Lake HPC Center (SSL-HPC) in Great Bay University. and the National Natural Science Foundation of China (Grant No. 62576076). 

\bigskip

\bibliography{aaai2026}

\end{document}